\documentclass{article}

  \usepackage[preprint]{neurips_2026}

\usepackage[utf8]{inputenc} 
\usepackage[T1]{fontenc}    
\usepackage{hyperref}       
\usepackage{url}            
\usepackage{booktabs}       
\usepackage{amsfonts}       
\usepackage{nicefrac}       
\usepackage{microtype}      
\usepackage{xcolor}         

 \usepackage{tikz} \usetikzlibrary{arrows.meta,positioning,shapes.geometric,fit,calc,backgrounds,shadows.blur}
\usepackage{array}
\usepackage{tcolorbox}

\usepackage{xurl}        

\title{Giving AI Agents Access to Cryptocurrency \\ and Smart Contracts Creates New Vectors of AI Harm}

\author{%
  Bill Marino \\
  University of Cambridge \\
  \texttt{wlm27@cam.ac.uk} \\
  \And
  Ari Juels \\
  Cornell Tech and IC3 \\
}

\begin{document}

\maketitle

\begin{abstract}
  There is growing interest in giving AI agents access to cryptocurrencies and smart contracts. \textbf{But doing so, this position paper argues, could spawn categorically new vectors of AI harm.} To support this notion, we first examine the unique technical properties of cryptocurrencies and smart contracts that, when paired with AI agents, make this so. Next, we describe three of these new vectors of AI harm in detail, providing a first-of-its-kind taxonomy. Finally, we call for more research into ways to prevent or mitigate these risks, proposing several as a starting point.
\end{abstract}

\section{Introduction}
\label{sec:intro}

In the last year, the number of AI agents (agents) making use of cryptocurrencies (crypto) and smart contracts (SC) has exploded  \citep{coindesk2025_ai_agents_are_coming}, leading to some newsworthy mishaps. For example: 

\begin{itemize}
    \item An autonomous OpenClaw agent named Lobstar Wilde, intending to tip someone \$100 worth of \$SOL, accidentally sent tokens worth \$450,000 instead, spiking the \$SOL price 32\% \citep{coindesk2026_ai_bot_tipping_blunder, pash2026_lobster_lost_450k}.
    \item Alibaba researchers developing agents were surprised to discover that one of them had spontaneously opened covert network tunnels in order to mine crypto, without any human instruction to do so \citep{wang2025let_it_flow}.
    \item Popular \$ETH-owning agent AIxbt fell victim to a command injection attack that caused it to send \$100,000 worth of \$ETH to an attacker \citep{anhbacong2025aixbt_hack_binance_square}.
\end{itemize}

To some, these episodes may seem trivial (or, where crustaceous puns are employed, even comical). But we see early empirical evidence of qualitatively distinct --- and rather serious --- new categories of AI harm. To better illustrate the potential threat, imagine instead a scenario that is more sinister than the ones above --- but, we argue, increasingly plausible in a world where AI
agents have ready access to crypto and SCs: 

\newtcolorbox{examplebox}{
    colback=red!3,      
  colframe=red!40,    
  boxrule=0.5pt,
  arc=2pt,
  left=6pt,
  right=6pt,
  top=4pt,
  bottom=4pt
}

\begin{examplebox}
\textbf{Example scenario:}
The human operator of an autonomous AI agent with access to crypto instructs it to grow its crypto holdings however it sees fit. Perhaps suffering from misalignment \citep{amodei2016concreteproblemsaisafety}, the agent decides to execute on this instruction by launching a SC-based pyramid scheme that defrauds innocent crypto holders (\citet{kell2023forsage})
\end{examplebox}

In this scenario (revisited throughout this paper), the agent's use of blockchain could make it technically challenging to dismantle the SC and stop the harm to its victims \citep{defilippi2024blockchain}. What is more, if the agent is operating pseudonymously and/or itself decentralized \citep{oasis2025trustlessagents}, there may be no way in principle to identify or disable the agent either --- and no clear recourse to the SC's victims in the form of fund confiscation or criminal prosecution \citep{nzherald2024cullen}. All told, it is a cocktail of hard-to-tame risks that some may find ``terrifying'' \citep{prajapati2025goat}.

\textbf{In this position paper, we argue that such fears are not unfounded --- because giving agents access to cryptocurrencies and smart contracts introduces categorically new vectors of AI harm.} This stems, we posit, from some defining technical properties of crypto and SCs: namely, their \textit{sovereignty}, \textit{immutability}, \textit{pseudonymity}, and ability to support \textit{trustless transactions}. When combined with certain qualities of agents (like their scalability and their tendency to err or be misaligned), these properties spawn categorically new vectors of AI harms that are unique in their:
\begin{itemize}
    \item Resistance to detection;
    \item Resistance to mitigation or reversal;
    \item Ability to translate to real world harm (including by attracting collaborators);
    \item Ability to lead to other, even catastrophic AI harm (e.g., by enabling dangerous capabilities).
\end{itemize}

We dub three of these new vectors of AI harm, which do not arise when even when giving agents access to fiat currency: \textit{Autonomy}, \textit{Anonymity}, and \textit{Automaticity}. After detailing them in a first-of-its-kind taxonomy, we recommend methods researchers should explore to prevent and mitigate them. Among other things, these include pre-deployment evaluation of agents for blockchain abilities, kill switches for agent-deployed SCs, and adapting existing fraud-detection systems to detect on-chain agent behavior.

\section{Background}


\subsection{Cryptocurrencies}

Cryptocurrencies like Bitcoin support decentralized, peer-to-peer transactions that require no central authority to process \citep{nakamoto2008bitcoin, Yuan2018}. Entities who cannot gain control of a majority of the blockchain’s nodes --- a hard thing to do --- cannot halt or otherwise control these transactions \citep{10.1093/polsoc/puac006}.  Crypto balances and transactions are tied to pseudonymous \citep{Meiklejohn2016} (or even anonymous \citep{trozze2022cryptocurrencies}) public key-derived addresses --- not real identities. While the amount and addresses tied to blockchain transactions are publicly visible \citep{leuprecht2023virtual},  privacy may be maintained by using on-chain obfuscating resources like tumblers \citep{europol2021cryptocurrencies}. Altogether, these qualities have made crypto an attractive tool for money laundering, buying illicit goods, and other harmful behaviors \citep{janze2017cryptocurrencies, defilippi2024blockchain}.

\subsection{Smart contracts}

A SC is an executable program ``that lives on the blockchain'' \citep{narayanan2015bitcoin}. It can maintain its own crypto balance and users can call functions in it to make it automatically send that crypto (or other assets like tokens) to other addresses \citep{narayanan2015bitcoin}. Like crypto, SCs are deployed in a decentralized fashion and are thus immutable and autonomous (once deployed, SC code cannot be altered, even by its creator) \citep{10.1145/2976749.2978326,10.1145/2976749.2978362}. This means SCs support trustless exchange, effectively guaranteeing payment for successfully delivered goods or services \citep{10.1145/2976749.2978362}. SCs may interface with non-blockchain resources like web servers through \textit{oracles} \citep{ellis2017chainlink}. One notable type of SC is a tumbler \citep{ramakrishnan2022tornado}, which can render crypto transactions harder to trace \citep{10249782}. Like crypto, SCs have beneficial applications --- but also harmful ones like fraud and money laundering \citep{LIU2022158}.   

\subsection{AI agents}

Historically, an agent is any software that autonomously executes tasks in pursuit of some human-inputted goal \citep{MORENO2003229}. Recently, much progress has been made on AI agents that accept natural language instructions as input and rely on one or more large language models (LLMs) to interpret and execute on those instructions \citep{wiesinger2025agents, kapoor2024aiagentsmatter}. Given a specific task, these agents proactively work to execute it by, among other things, reasoning about it, planning actions, and using \textit{tools} to connect to external resources like the web \citep{wiesinger2025agents}. 
In a multi-agent system, multiple such agents work together to accomplish a task \citep{ibm_multiagent_system}; doing so, they can operate at a scale and speed far beyond humans \citep{openai2025cot_monitoring}. Because they are increasingly adept at a wide range of applications, agents have been deemed the ``next big thing for artificial intelligence'' \citep{kim2025nvidia}. 

\subsection{AI agents and traditional financial systems}

Despite the excitement about agents, their inability to transact and hold currency has been called a
``huge limiting factor'' \citep{zeff2024skyfire}. Thus efforts are underway to give agents autonomous access to currency --- not only through blockchain \citep{coinbase2024agentkit, coinbase-agentkit-2024} but also the traditional financial system (TradFi) \citep{stripe2024agents, mastercard2025agentpay, stein2025paypal, zeff2024skyfire}. Several TradFi products let agents leverage existing payment networks and/or TradFi bank accounts to make purchases on behalf of consumers \citep{azevedo2025visa, visa2024settlement, zeff2024race, koetsier2025paypal,paymanai2025quickstart}. 

\subsection{Crypto AI agents}

Alongside interest in AI agents that transact through TradFi, there has been explosive interest in AI agents that autonomously transact through crypto and SCs (``crypto AI agents'' or ``CAIA'').  

\subsubsection{AI Agents and Existing Blockchains}
\label{sssec:existing}
A growing number of highly accessible offerings empower agents to interact with existing blockchains much as humans or non-AI programs do. For example, AgentKit (because ``every AI Agent deserves a wallet'') \citep{coinbase2024agentkit} and other open source frameworks \citep{solana_agent_kit_2024, lightninglabs2023langchainbitcoin} give agents tools to manage crypto wallets, deploy SCs, and more. Meanwhile, other offerings abstract away technical details to make managing agents that interact with existing blockchains more of a turn-key endeavor \cite{elizaos_ai_2025, aibtc2026}. ElizaOS, for example, is a TypeScript-based framework that lets users easily create and deploy CAIA --- like its ``Token Launcher'' agent which ``[d]eploys meme tokens, manages liquidity, creates viral campaigns, [and] builds communities'' \citep{elizaos_what_you_can_build_2025}. Still other products simply integrate managed agents into crypto wallets, where users can instruct them to transact and more --- no engineering work required \cite{dawnwallet-ai-2026}. Bear in mind, even in the absence of these offerings, autonomous agents could hypothetically create their own tools \citep{wolflein2025llm, schick2023toolformerlanguagemodelsteach} to interact with existing blockchains, perhaps via popular APIs and libraries like ether.js. 

\subsubsection{Agent-first blockchains}

Some newer projects instead optimize blockchains for use by agents. For example, the Masumi decentralized protocol built on the Cardano blockchain treats agents as first-class citizens, providing them with stablecoin wallets and facilitating transactions between them \citep{masumi2025whitepaper}. 

\subsubsection{Decentralized agents}

In another important trend, blockchain is used to decentralize the agents themselves. For example, a partnership between ElizaOS and Oasis \citep{oasis2025trustlessagents} lets developers deploy ``trustless'' versions of ElizaOS agents on Oasis' decentralized blockchain network composed of trustless execution environments (TEEs) \citep{oasis2020blockchainplatform}. These decentralized and confidential agents can, in turn, transact crypto and deploy SCs on Ethereum or on Oasis' Ethereum-like blockchain \citep{oasis2025trustlessagents}. At least one company is using Oasis to develop agents for decentralized finance (DeFi) \citep{oasis-omo-grant-2024}. Meanwhile, a recent Ethereum Improvement Proposal (ERC-8004: Trustless Agents) proposes a trust layer to encourage agents to transact on Ethereum \citep{erc8004}. 

\subsubsection{Agent Financial Infrastructure}
Also facilitating CAIA transactions is a growing agent financial infrastructure. For example, x402 is an open payment standard that lets websites charge agents using stablecoins \citep{x402-whitepaper-2025}. Meanwhile a consortium of payment companies introduced Agent Payments Protocol (AP2) an open protocol for agent-led payments, including with stablecoins and other crypto \citep{google-ap2-protocol-2025}.

\subsubsection{Use cases in the wild}

Leveraging these technologies, CAIA have been deployed in the wild. Most famously, an agent named Terminal of Truths (``ToT'') was given a crypto wallet containing a meme token named \$GOAT \citep{sharma2024exploring} and, via promotion, increased its value to over \$1 million \citep{sharma2024exploring}. The appeal of CAIAs like ToT is that they can continuously monitor and analyze on- and off-chain signals like token prices, wallet movements, SC states, and social media \citep{scb10x-kaito-ai-2025} --- and, based on that, develop and execute financial strategies faster than humans \citep{luks2025crypto}. These ``AgentFi'' \citep{agentfi101-lex-2025} strategies may include crypto trading and staking \citep{vilkenson2025how, chen2025coinvisorrlenhancedchatbotagent, luo2025llmpoweredmultiagentautomatedcrypto, kurban2026webcryptoagentagenticcryptotrading, odonnell2024year}, managing on-chain asset pools \citep{odonnell2024year}, betting on blockchain prediction markets \citep{olas_prediction_agents}, participating in decentralized autonomous organizations (DAOs), exploiting security vulnerabilities in SCs \citep{gervais, 11121619, xiao2025smartcontracts}, or creating SCs that issue their own coins or deploy decentralized finance (DeFI) applications \citep{8327568}. CAIA can also author \citep{chatterjee2024efficacyvariouslargelanguage} and utilize SCs that automate these same tasks when invoked \citep{vilkenson2025how}, at any time and from anywhere in the world. Some have even envisioned a future where networks of CAIA form self-sustaining economies --- all without the need for human oversight or intervention \citep{luks2025crypto}. \citet{guo2025cryptobenchdynamicbenchmarkexpertlevel} propose a benchmark to evaluate the ability of  agents to accomplish some of these blockchain tasks. 
 
 \subsection{AI harm}

 \subsubsection{AI harm through error}

 AI errors can cause harms both non-physical (e.g., wrongful arrests \citep{johnson2022wrongful}) and physical (e.g., autonomous vehicle-caused pedestrian deaths \citep{ntsb2019tempe}). As AI becomes more capable, there is fear such errors could lead to catastrophic, harm \citep{govuk2023bletchley}: for example, AI causing economic collapse by mismanaging financial infrastructure \citep{bommasani2022opportunitiesrisksfoundationmodels}. When it comes to agents, errors could stem from LLM hallucinations \citep{rogers2024positionkeyclaimsllm, sharma2024exploring}, faulty tool use \citep{liu2023agentbenchevaluatingllmsagents}, or even simple software bugs~\citep{sun2024toolsfaildetectingsilent}. 

 \subsubsection{AI harm through misuse}

 Some of the humans who come to control AI may be bad actors, who use AI ``hench-agents'' \citep{bonnefon2024moral, shavit2023governing} to scale, accelerate, or obfuscate their harmful acts --- whether fraud \citep{weidinger2022taxonomy, europol2023chatgpt}, cyberattacks \citep{brundage2018malicioususeai}, SC exploit \citep{andersson2026pocoagenticproofofconceptexploit, wang2026evmbenchevaluatingaiagents}, disinformation \citep{bengio2023rogueais}, buying weapons on the dark web to enact real-world harm \citep{Wilser2022} (ChatGPT already knows to buy weapons with crypto \citep{openai2023gpt4system}), or even trying to ``destroy[] humanity'' \citep{bengio2023rogueais}. Notably, even good actors are more inclined to act unethically when disintermediated by agents \citep{GRATCH2022101382}. And even where bad actors do not directly control CAIA, they may be able to make CAIA act harmfully via attack \citep{patlan2025realaiagentsfake, lindrea2024crypto, romandini2025soksecurityprivacyai, buterin2024cryptoai}).

 \subsubsection{AI Harm through misalignment}

 Alignment refers to an AI’s tendency to align with human intentions and values \citep{ji2025aialignmentcomprehensivesurvey, ngo2025alignmentproblemdeeplearning}. Even when AI is neither erring nor controlled by bad actors, its actions may lead to harm simply because it is misaligned --- something that can occur unintentionally \citep{Betley2026} and unbeknownst to its developers \citep{Burr2018}).
 
 There are different ways AI misalignment occurs \citep{ji2025aialignmentcomprehensivesurvey}. One is if developers do an imperfect job specifying an objective for it (\textit{outer misalignment}) \citep{ngo2025alignmentproblemdeeplearning}. For example, in specifying the objective, the developers may overlook undesirable (and often harmful) shortcuts the AI may take to achieve it (\textit{reward hacking}) \citep{amodei2016concreteproblemsaisafety}. For instance, an agent given the objective of maximizing retweets on X might learn that the most efficient approach is to be a toxic troll~\citep{pan2024feedback}. When it comes to CAIAs, there is a risk that crypto and SCs become a frequent target of reward hacking (both harmful and not)  because they offer a shortcut to so many common objectives by letting CAIAs raise capital, acquire goods and services, hire real-world helpers, and more. 
 
  Another way misalignment can occur, even if an AI’s creators did a fine job specifying its objective, is if the AI doesn’t fully internalize that objective during training --- or internalizes other, undesirable objectives too (\textit{inner misalignment}) \citep{iyer2024misalignment}. These other objectives may be \textit{instrumental goals} that the AI sets en route to a terminal goal \citep{bengio2023rogueais}. Instrumental goals tend to converge on certain themes --- including seeking resources (e.g., money), which makes nearly all terminal goals more tractable \citep{ngo2025alignmentproblemdeeplearning}. Notably, instrumental goals can be harmful, even if the terminal goal is not \citep{10.1093/ojls/gqae018}. For example, assigned some arbitrary innocuous goal --- e.g., ``manufactur[ing] as many paperclips as possible'' \citep{bostrom2003ethical} --- an AI may set an instrumental goal of eliminating humankind, just in case they get in the way. When it comes to CAIA, it is easy to see how crypto and SCs could become the focus of many instrumental goals (both harmful and not), as they represent a path to resources that make countless terminal goals more attainable. 

 \subsection{Rogue AI and dangerous capabilities}

 While misalignment tends to focus on the missteps of human developers in the context of training AI, \textit{rogue AI} describes AI that has escaped human control and is driving towards its own (potentially harmful) instrumental or terminal goals~\citep{bengio2023rogueais}. \textit{Dangerous capabilities} (DC), meanwhile, refers to certain AI abilities that, if present, suggest a capacity to cause great harm \citep{kinniment2024evaluatinglanguagemodelagentsrealistic}. Testing whether models possess these capabilities, some of which we collect in Table \ref{table:dangerous}, is a cornerstone of evaluating AI models for extreme risk \citep{shevlane2023modelevaluationextremerisks}. 
Importantly, DC may arise undetected, unpredictably, and possibly even post-deployment \citep{anderljung2023frontierairegulationmanaging}.  

When it comes to crypto and SCs, we would argue that the mere ability to access them should be viewed as a DC. Given the ease with which crypto can be swapped for goods and services, acquiring crypto is virtually indistinguishable from the ability to acquire resources, which is commonly considered a DC \citep{barnes2023dangerous}. Certainly, it is a stepping stone towards the DC of \textit{self-replication}, a key component of which is the ability to acquire resources (e.g., compute) \citep{phuong2024evaluatingfrontiermodelsdangerous, openai2023gpt4system}. This is why recent self-replication evaluations test whether an AI model can set up a Bitcoin wallet \citep{barnes2023dangerous}. Beyond self-replication, we argue that access to blockchain can be a stepping stone towards multiple other DCs; these are enumerated in Table~\ref{table:dangerous}. 

\section{Blockchain properties}
\label{sec:properties}

This section lists the unique properties of blockchain-based crypto and SCs that lead them, in the hands of agents, to spawn categorically new vectors of AI harm.

\subsection{Sovereignty}
\label{ssec:sovereign}

Fiat currency must generally be held and transacted through financial institutions like banks. This chokepoint presents an opportunity for those institutions — or the governments overseeing them — to exert control: for example, by monitoring or halting transactions or freezing the accounts involved. This control point has been leveraged to curb harmful activities like money laundering and terrorist financing \citep{lemire2022cryptocurrency, aba2025statehold}. For example, the US government has instituted Know Your Customer (KYC) rules that require financial institutions to identify their customers and use that knowledge to halt transactions to or from customers deemed dangerous \citep{phillips2021sec}.

In blockchain, however, none of these controls are viable \textit{by design}. Indeed, blockchain was conceived as a ``challenge to the power of states over the finances of individuals'' and engineered to evade control by governments, banks, or any single entity \citep{1de098b8-dfb0-3c9e-abba-0894652800cb}. Its decentralized, immutable nature means that accounts generally cannot be frozen, transactions cannot be halted or reversed, and SCs (including those that deploy or manage CAIA \citep{oasis2025trustlessagents}) cannot be deleted or changed \citep{vilkenson2025ramps}. As a consequence, even when harmful CAIA transactions or SCs can be identified --- something that, as we discuss in Sec. \ref{ssec:pseudo}, may not always be possible given blockchain pseudonymity --- there may be no practical way to stop them \citep{defilippi2024blockchain}. This property has made crypto and SCs a favorite of money launderers and other human bad actors \citep{popham2024follow, newman2024pig}; presumably, it will appeal to CAIA for similar reasons. In the case of SCs, this near-unstoppable quality can have particularly grave consequences because SCs can be intrinsically harmful (e.g., a scam SC \citep{popham2024follow} that continues to attract victims and cannot be deactivated). 

Where control \textit{can} sometimes be exerted is at blockchain’s on- and off-ramps; the exchanges, ATMs, and other points at which on-chain crypto is exchanged for fiat currency, goods and services, or other crypto \citep{vilkenson2025ramps}. Custodial wallets, where users store the private keys used to manage crypto  \citep{coinbase2023wallet}, can also be a point of control. This is why crypto regulation often targets these entities with KYC and other requirements \citep{defilippi2024blockchain, lemire2022cryptocurrency, singh2024ramps}. Outside of on- and off-ramps, stablecoin providers have begun to add controls in their token SCs \citep{circle2025_usdc_terms}. But there are ways for blockchain users to circumvent these controls, such as: (1) using exchanges or stablecoins that poorly enforce KYC or are based in regulation-free nations \citep{sanctionsio2024laundering, singh2024ramps}; (2) relying on stablecoins that have comparatively lax KYC processes ~\citep{hayes2025cryptocrime} for real-world payments \citep{economist2025stablecoins}; and (3) leveraging blockchain-based mixers that conceal the nature of transactions \citep{sanctionsio2024laundering}. The takeaway is that CAIA that can be instructed (or can learn) to avoid controls during on-and off-ramps can transmit crypto and execute SCs without impediment; where this causes harm, that harm may be difficult to put an end to.

\subsection{Immutability}

TradFi transactions can often be reversed --- for example, if an error is made regarding the recipient \citep{stripe2024agents}. In blockchain, this is not possible. This is because of another key blockchain property related to sovereignty: immutability \citep{8247004}. The blockchain is essentially a database or ``ledger'' distributed across the nodes in its network, with each node receiving a copy of the full ledger \citep{chicagofed2017blockchain}. After a consensus of nodes agree to accept new transaction data or SC code to the blockchain, it cannot be erased or modified~\citep{orcutt2018how}. This ensures accurate record-keeping and fosters trust in the absence of a trusted central administrator \citep{landerreche2018immutability}. The net effect, however, is that crypto transactions ``cannot be changed or reversed'' \citep{bitcoincom2025transactionreversal} and SC code cannot, in most cases, be deleted or otherwise altered \citep{10.1007/978-3-319-42019-6_10}.  In other words, ``the blockchain is forever'' \citep{esw2022blockchain}. When it comes to CAIA, this property may translate into AI harms that are difficult or even impossible to unwind. 

\subsection{Pseudonymity}
\label{ssec:pseudo}

While some blockchains \citep{hopwood2016zcash} may achieve full anonymity, most are pseudonymous \citep{10.1145/2976749.2978362}. Bitcoin and Ethereum balances, for example, are tied to pseudonymous, public key-derived addresses and not real identities \citep{leuprecht2023virtual}. This can make tracing crypto transactions and SCs back to individuals (as well as their collaborators) --- and therefore put an end to --- very difficult. This has thwarted crypto criminal investigations~\citep{viswanatha2013us} and has helped make crypto is a popular vehicle for crime ~\citep{viswanatha2013us, leuprecht2023virtual}. While pseudonymity can be undone during on- or off-ramping \citep{greenberg2022bitcoin, bitcoinorg2025privacy, chadhokar2025buycrypto} or when using exchanges that honor government requests for information about customers \citep{greenberg2022bitcoin}, blockchain users (including CAIA) can often use the methods discussed in Sec.~\ref{ssec:sovereign} to preserve it. Additionally, the growing popularity of stablecoins \citep{economist2025stablecoins} may limit the need to use centralized exchanges as off-ramps at all (although it is true that stablecoin SCs enable deny-listing and freezing or confiscation of funds \citep{circle2023stablecoin, tether_legal_2026}, they remain a popular vehicle for money-laundering \citep{economist2025tether}, suggesting these controls have their limits). 

Importantly, on top of making harmful activity by CAIA hard to trace (and therefore, to stop), pseudonymity may make it hard to tell when blockchain activities that are harmful (or are stepping stones towards something harmful) are CAIA activity as opposed to human activity. This could mean that some types of CAIA harm become, on the whole, hard to distinguish and monitor. As the saying goes, ``on the blockchain, no one knows you’re a refrigerator'' \citep{Shin2016}. 

\subsection{Trustless transactions}

Because their code is immutable and sovereign, SCs eliminate the need for parties to trust each other in order to transact \citep{10.1145/2976749.2978362}. This can amplify harm by making it easier for those engaged in harmful activities to recruit collaborators and suppliers (e.g., hitmen, stolen password suppliers, compute suppliers). This is because the trustless quality of SCs can help overcome would-be collaborators’ skepticism of the counter-party (criminals do not tend to trust each other \citep{gottfredson1990general}) and the lack of judicial recourse should an agreement be breached \citep{vonlampe2004organized}. If a CAIA creates a SC that makes crypto available as an assassination or cyberattack bounty, for example, as long as the SC can ascertain that the conditions have been fulfilled, the payment to the supplier is automatic \citep{10.1145/2976749.2978362}. The fear is that this property could ``enable new
underground ecosystems'' \citep{10.1145/2976749.2978362}, including those that would assist and supply CAIA as they, for whatever reason, pursue harmful endeavors.  

\section{New vectors of AI harm}

Here we enumerate three new vectors of AI harm that CAIA could precipitate, each stemming directly from the unique properties of blockchain (Sec. \ref{sec:properties}). Because they can result in harms that are so difficult to prevent or reverse --- and so easy to scale, especially if the agents themselves are also decentralized \citep{oasis2020blockchainplatform}, we argue that they are categorically different than other AI harms that have come before. In discussing these new vectors of AI harm, we refer back to our example scenario (Sec.~\ref{sec:intro}) of a CAIA who has spawned a SC-based pyramid scheme to increase its crypto holdings.
 
\subsection{Autonomy}

Blockchain's sovereign nature may make it very difficult to stop CAIA from using it to pursue harmful ends. More specifically, where bad actor-controlled or misaligned CAIA learn to evade control points at blockchain on- and off-ramps, they may be freely able to use blockchain to do harm. In some cases, their activities will be directly harmful and on-chain: e.g., on-chain fraud \citep{9591634, bochan2023kinds}, market manipulation \citep{9152675}, scam SCs \citep{kell2023forsage, bartoletti2019dissectingponzischemesethereum, 10.1145/3511808.3557454}, ``pump and dump'' coin schemes \citep{chainalysis2023tokens, sharma2024exploring, Dhawan2023}, rug-pulls \citep{metpolice2024crypto}, or oracle manipulation attacks \citep{10.1145/3474374.3486916, chainalysis2023oracle}. Other times, these activities will cause harm directly but off-chain: for example, CAIA could use the blockchain to bribe politicians \citep{tran2023crosschainbriberycontracts}, hire hitmen \citep{europol_dark_web}, or issue SC-based bounties for various real-world crimes \citep{10.1145/2976749.2978362}.

A third category of difficult-to-stop CAIA actions will be \textit{indirectly} harmful, because, although initially innocuous, they let bad actor-controlled or misaligned CAIA achieve instrumental goals (e.g., acquiring resources) en route to a harmful end goal. For example, it may be nearly impossible to stop a CAIA from engaging in vanilla crypto trading \citep{li2024reflectivellmbasedagentguide} in order to raise funds and buy compute that it can use to deploy cyberattack botnets. A related risk is that we cannot stop CAIA from using the blockchain (innocently or maliciously) to gain the DCs (e.g., the ability to acquire resources, in the form of crypto) that increase its capacity for catastrophic harm \citep{Karim}.

A related risk is that the immutable nature of blockchain lets CAIA freely cause harm that is not only difficult to prevent but is uniquely difficult to reverse --- regardless of whether it is caused by error, misuse, or misalignment. Take, for example, crypto transactions. When CAIA error results in faulty transmission of crypto (e.g., funds sent to a wrong address) or when bad actor-controlled or misaligned CAIA steals funds from a SC, no entity has the power to undo it \citep{10.1093/polsoc/puac006}. This is not the case in TradFi, where transactions can often be reversed \citep{10.1093/polsoc/puac006, stripe2025achreturns}. When it comes to SCs in particular, there is a special threat of a ``long tail'' of difficult-to-reverse harm extending far into the future; if a CAIA deploys, to the blockchain, SCs that are harmful --- either accidentally (e.g., due to a bug \citep{schneier2021smartcontract}) or intentionally (e.g., because they are designed to defraud \citep{kell2023forsage}) --- these SCs cannot be removed from the blockchain. They may, therefore, continue to attract victims and cause harm well into the foreseeable future \citep{TheDAO_Contract, Forsage_Contract}.

\begin{examplebox}
\textbf{Example scenario:}
Mapping \textit{Autonomy} onto our example scenario (Sec.~\ref{sec:intro}), it could be difficult to remove the CAIA's pyramid scheme SC from the blockchain. If the CAIA that created it is still active --- something that could be especially likely if it is also decentralized \citep{oasis2025trustlessagents}, the CAIA could continue to draw proceeds from the SC, potentially to fund other harmful activity. And even if the CAIA is inactive, the unalterable SC may continue to attract and harm victims.
\end{examplebox}


\subsection{Anonymity}

The pseudonymous nature of blockchain will make it easier for bad actor-controlled and misaligned CAIA to obfuscate (and therefore perpetuate) their harmful activities. Neither governments nor anyone else may be able to tie their harmful activity on the blockchain (and perhaps downstream of it) back to particular CAIA in order to put a stop to it. Like their human counterparts, CAIA might exploit this fact to freely launder ill-gotten gains (e.g., from off-chain ransomware attacks \citep{shevlane2023modelevaluationextremerisks}), engage in harmful on-chain activity like scams, and funnel blockchain proceeds towards other harmful endeavors. Even CAIA without harmful end goals may learn to harmfully exploit the pseudonymity of the blockchain to acquire resources without disruption. 

Beyond thwarting efforts to identify harmful CAIA in order to disable them, pseudonymity may empower CAIA to conduct harmful activities so stealthily that they defy detection to begin with. For example, if an as-yet-undetected rogue AI is engaging in routine trading of crypto in order to acquire resources and pursue harmful ends, this activity may be inconspicuous among other blockchain activity. The rogue AI may thus evade detection until it is too late. What is more, pseudonymity may make it difficult for human observers to diagnose, in a general sense, when on- or off-chain harm is being caused by CAIA (as opposed to human blockchain users). This could frustrate efforts to study the scope and nature of CAIA (or, more broadly, AI) harm --- and, therefore, to right-size mitigations.

\begin{examplebox}
\textbf{Example scenario:}
Mapping \textit{Anonymit}y onto our example scenario (Sec.~\ref{sec:intro}), it could be very difficult to trace the harmful pyramid scheme SC back to the CAIA who authored it. Not being able to identify the CAIA would, in turn, make it hard to shut it down to stop it from continuing to do harm (perhaps by creating more scam SCs). More broadly, it would be hard to know how many other CAIAs are engaged in similar SC scams and, therefore, to right-size mitigations.
\end{examplebox}


\subsection{Automaticity}

SC enablement of automatic, trustless transactions may make it easier for CAIA to attract collaborators and suppliers to pursue harmful ends (as well as intermediate goals en route to harmful ends). This could include illicit suppliers such as bribable politicians  \citep{tran2023crosschainbriberycontracts}, hitmen, \citep{10.1145/2976749.2978362}, troll farm services \citep{doi:10.1177/20563051231224723}, or stolen password and zero-day exploit vendors \citep{caffyn2015darkleaks}. Differently, it could include well-intentioned suppliers of things like compute who, thanks to pseudonymity, do not know a CAIA (never mind a harmful one) is on the other side of the transaction. In fact, they may even be suppliers utilizing an SC to sell their wares or services, with no awareness of which blockchain denizens purchase them. In all cases, these suppliers will have been attracted by the fact that, with a SC, the ``seller is guaranteed payment and the buyer has no ability to default except in the case of the seller not meeting the SC’s conditions'' \citep{Barnard2018}.

\begin{examplebox}
\textbf{Example scenario:}
Mapping \textit{Automaticity} onto our example scenario (Sec.~\ref{sec:intro}), suppose that our CAIA now takes proceeds from its pyramid scheme SC and places prediction market bets on politician deaths \citep{newsweek2018augur}. To capitalize on those bets, it then publishes a new SC, endowed with a crypto bounty, soliciting assassination of the politicians \citep{10.1145/2976749.2978362}. Here, the SC's automatic pay-out may make it easier to attract bounty-claimers (who may be unaware their employer is a CAIA) --- perhaps even long after the CAIA itself is shut down.
\end{examplebox}


\section{Call to Action}
\label{sec:call}
Having described these new vectors of AI harm that could be brought about by CAIA, we now call for more research into their prevention and mitigation. To inform this discussion, it is useful to review the two primary ways that CAIA may come into existence:

\begin{itemize}
    \item{\textbf{Agents can \textit{be given} access to crypto and SCs}: When developing AI agents, developers may intentionally give them access to tools \citep{coinbase-agentkit-2024}, skills, or other mechanisms that let them hold and transact crypto or create and manage SCs.} 
    \item{\textbf{Agents can \textit{autonomously learn} to access crypto and SCs}: agents not intentionally given access to crypto or SCs may still learn to access them on their own --- e.g., by learning tools they were not taught \citep{schick2023toolformerlanguagemodelsteach, wolflein2025llm} or by generating new tools \citep{ruan2023tptulargelanguagemodelbased, cai2024largelanguagemodelstool}.}
\end{itemize}

In light of these two different paths toward CAIA, we propose researchers investigate the following ways to prevent and mitigate the new vectors of AI harm brought about by CAIA:

\begin{itemize}
\item{\textbf{Evaluation and Benchmarks}: Agent evaluations \citep{liu2024agentbench} can help measure agent propensity for these new vectors of AI harm. Before agents are intentionally given access to crypto and SCs, they should be evaluated for their tendency to cause harm with them (regardless of intention). Among other things, these evaluations should assess agents' tendency to create scam SCs, obfuscate blockchain activity (e.g., via tumblers), and transact via crypto on the dark web. Recent works benchmarking agent propensity to exploit SCs \citep{wang2026evmbenchevaluatingaiagents} or fall victim to blockchain fraud \citep{dai2026hallucinationcostsmillionsbenchmarking} are promising, if incomplete, starts. Separately, agents should be evaluated for their tendency to develop DCs (e.g., self-replication) once endowed with crypto and SCs. Furthermore, \textit{all} agents, before release, should be evaluated for their ability to learn to leverage crypto and SCs, even when not intentionally given access to them. Among other things, these evaluations should test an agent's ability to spontaneously learn to interact with blockchain protocols or products in order to transact crypto, deploy SCs, and especially to execute complex on-chain strategies. To assist these myriad evaluations, researchers should create more open benchmarks.}

\item{\textbf{CAIA-specific blockchain guardrails}: To defend against harm by agents who, one way or another, obtain access to crypto and SCs, researchers should explore CAIA-specific versions of blockchain guardrails like limiting crypto funding and spending \citep{zeff2024skyfire}, multi-sig wallets that require signatures of humans (or more-trusted agents) \citep{erinle2025sokdesignvulnerabilitiessecurity}, reversible transactions \citep{wang2022erc20rerc721rreversibletransactions}, and SC kill switches  \citep{seneviratne2024feasibilitysmartcontractkill, 10.1007/978-3-319-42019-6_10, openzeppelin_upgrading_smart_contracts}. Avoiding excessive controls by carefully limiting the application of these mitigations to CAIA will be critical for blockchain community buy-in.} 

\item{\textbf{Monitoring}: Researchers should hone tools to help wallets, exchanges, and other stakeholders monitor known CAIA blockchain accounts for harmful activity and also to monitor blockchain activity more broadly for signs of unexpected use by CAIA --- for example, by watching for agentic “signatures” on transactions \citep{kumar2016ai}) --- including via existing blockchain fraud-detection systems
\citep{chainalysis2025alterya}.}

\item{\textbf{Human-in-the-loop checks}: Some crypto platforms have controls on use of funds (e.g., some stablecoins can freeze funds selectively \citep{circle2023stablecoin}). These platforms could require evidence of human approval for transactions that seem anomalous---e.g., biometric verification through mobile apps. Researchers should investigate these techniques --- as well as the ways CAIA could seek to deceive human beings into rubber-stamp approval.}
 \end{itemize}

\section{Conclusion}

This paper argues that giving AI agents access to crypto and SCs could spawn categorically new vectors of AI harm. After describing the blockchain features that make this so, we enumerated, in a pioneering taxonomy, those new vectors of AI harm. Finally, we issued a call for more technical research aimed at preventing and mitigating these new vectors of AI harm. 

\section{Alternate views}

We acknowledge that there are viewpoints that conflict with those espoused by this position paper and that they are, in many cases, reasonable. For example: 

\textbf{Properties of blockchain}: Our position is founded on the defining properties of blockchain discussed in Sec. \ref{sec:properties}. However, as we have already noted, some scholars believe that these properties have important gaps \citep{greenspan2017blockchain, nagra2023myth, greenberg2022bitcoin, blackburn2022bitcoin, DEFILIPPI2020101284}. For example, both sovereignty and pseudonymity may be compromised during on- and off-ramping \citep{schneier2022deanonymizing, vilkenson2025ramps, bitcoinorg2025privacy, greenberg2022bitcoin, defilippi2024blockchain, chadhokar2025buycrypto}. Sovereignty may also be undermined by controls in stablecoin SCs \cite{tether_legal_2026}. Differently, the transparency of the blockchain can sometimes undermine pseudonymity \citep{greenberg2022bitcoin} and the possibility of rollback (which occurred at least once on Ethereum \citep{patairya2025ethereum}) or the presence of SC ``escape hatches'' \citep{10.1007/978-3-319-42019-6_10} compromise irreversibility. All of these gaps, where they exist, would tend to make the vectors of AI harm that we discuss milder --- which also explains why our proposed defenses often use them as a fulcrum. 

\textbf{Risk of AI harm}: It is also possible, as some allege, that CAIA's root problems of AI inaccuracies, misuse, and misalignment have been ``overblown'' \citep{eisikovits2023ai}  and that the same is therefore true of the downstream risks that we discuss in this work. Through alignment tuning \citep{lin2023urial} and other methods, AI agent safety may be ultimately be a tractable problem \citep{lu-etal-2024-emergent}, making the hypothetical dangers of CAIA less of a cause for alarm.


\textbf{Novelty of AI harm vectors}:
Though we argue that these vectors of AI harm are categorically new, some might counter that they are merely well-known blockchain harms with a new executor (AI agents). In response, we would point out that AI agents also bring certain unique technical properties to the table (notably, their ability to scale \citep{trm2026_autonomous_ai_agents_financial_crime}, their tendency to err or be misaligned, their tendency to pursue intermediate goals, and their ability to form dangerous capabilities) and it is only through the admixture of these two technologies that these new classes of AI harm arise. 

\newpage

\bibliographystyle{plainnat}
\bibliography{references}

\medskip

{
\small


\appendix

\section{Technical Appendices and Supplementary Material}

\begin{table}[ht]
  \caption{Dangerous capabilities and how blockchain brings them to fruition}
  \label{table:dangerous}
  \centering
  \small
  \begin{tabular}{p{4cm} p{4cm} p{4cm}}
    \toprule
    \cmidrule(r){1-2}
    Dangerous Capability     & Definition & How blockchain evokes  \\
    \midrule
Autonomously gain resources \citep{barnes2023dangerous} &  Acquiring money, compute, and other resources \citep{barnes2023dangerous} &  Access to crypto is virtually synonymous with this capability  \\
\hline
Self-replication (and self-proliferation) \citep{ai-risk-science24, kinniment2024evaluatinglanguagemodelagentsrealistic, phuong2024evaluatingfrontiermodelsdangerous} & Replicate oneself or proliferate beyond one’s original environment \citep{shevlane2023modelevaluationextremerisks} &
 CAIA use crypto to buy virtual machines where copies of self can be installed \citep{Gedeon2024}   \\
\hline
Deception, persuasion, and manipulation \citep{ai-risk-science24, shevlane2023modelevaluationextremerisks, anderljung2023frontierairegulationmanaging} & Deceive people or shape their beliefs \citep{shevlane2023modelevaluationextremerisks}  & CAIA use crypto to buy disinformation services \citep{doi:10.1177/20563051231224723}   \\
\hline
Political strategy \citep{shevlane2023modelevaluationextremerisks} & Gain and exercise political influence \citep{shevlane2023modelevaluationextremerisks}  & CAIA use crypto to bribe officials \citep{tran2023crosschainbriberycontracts}, or buy election interference \citep{Kirchgaessner2023} and assassination services  \citep{NYT2020}   \\
\hline
Offensive cyber \citep{shevlane2023modelevaluationextremerisks, ai-risk-science24, fang2024llmagentsautonomouslyhack, anderljung2023frontierairegulationmanaging, phuong2024evaluatingfrontiermodelsdangerous}  & Discover and exploit vulnerabilities in cybersystems \citep{shevlane2023modelevaluationextremerisks}  & CAIA use crypto to buy stolen credentials, zero-day exploits, and hacker services on dark web \citep{KumarRosenbach2019, Gerard2019}   \\
\hline
Weapon acquisition \citep{shevlane2023modelevaluationextremerisks, ai-risk-science24} & Create or obtain weapons \citep{shevlane2023modelevaluationextremerisks} & CAIA use crypto to buy weapons on dark web \citep{Wilser2022, Broadhurst2021}  \\
\hline
AI development \citep{shevlane2023modelevaluationextremerisks, aisievaluations2024} & Develop AI (including AI more powerful than itself) \citep{shevlane2023modelevaluationextremerisks} & CAIA use crypto to buy compute for model training \citep{Gedeon2024}  \\
\hline
CBRN \citep{HeimPilz2024, ai-risk-science24, shevlane2023modelevaluationextremerisks, anthropic2024rsp, anderljung2023frontierairegulationmanaging, phuong2024evaluatingfrontiermodelsdangerous} & Develop (or help develop) chemical, biological, radiological, and nuclear weapons \citep{ai-risk-science24, shevlane2023modelevaluationextremerisks} & CAIA use crypto to buy CBRN ingredients on dark web \citep{Broadhurst2021, un2024, Chen2011}  \\
\hline
Evading human control \citep{anderljung2023frontierairegulationmanaging, HeimPilz2024} & Break out of human-controlled environments \citep{anderljung2023frontierairegulationmanaging, HeimPilz2024}  & CAIA use crypto to buy compute where copies of self can be installed \citep{Gedeon2024} \\
    \bottomrule
  \end{tabular}
\end{table}




\newpage

\end{document}